%% file: main.tex
\documentclass[10pt,twocolumn,letterpaper]{article}

\usepackage[pagenumbers]{cvpr} 

\input{preamble}

\definecolor{cvprblue}{rgb}{0.21,0.49,0.74}
\usepackage[pagebackref,breaklinks,colorlinks,allcolors=cvprblue]{hyperref}

\def\paperID{} 
\def\confName{CVPR}
\def\confYear{2026}

\title{RoboTAG: End-to-end Robot Configuration Estimation via \\ Topological Alignment Graph}

\author{
Yifan Liu$^{1,2}$\quad Fangneng Zhan$^{2,3}$\quad Wanhua Li$^{2,4}$\quad Haowen Sun$^{1}$\\
Katerina Fragkiadaki$^{5}$\quad Hanspeter Pfister$^{2\dagger}$\\
\small
$^{1}$Tsinghua University\quad
$^{2}$Harvard University\quad
$^{3}$Massachusetts Institute of Technology \\
\small
$^{4}$Nanyang Technological University\quad
$^{5}$Carnegie Mellon University\\
{\tt\small liuyifan22@mails.tsinghua.edu.cn,\quad pfister@seas.harvard.edu}
}

\begin{document}
\maketitle
\input{sec/0_abstract}    
\input{sec/1_intro}
\input{sec/2_related}

\input{sec/3_method}

\input{sec/4_experiments}
\input{sec/5_conclusion}

{
    \small
    \bibliographystyle{ieeenat_fullname}
    \bibliography{main}
}

\input{sec/X_suppl}

\end{document}

%% file: preamble.tex
\usepackage[accsupp]{axessibility}  

\definecolor{deepgray}{rgb}{0.3,0.3,0.3}
\newcommand{\growth}[1]{\textcolor{deepgray}{\footnotesize (#1)}}

\usepackage{xcolor}

\usepackage{algorithm}
\usepackage{algorithmic}
\usepackage{multirow}
\usepackage{makecell}



%% file: sec/0_abstract.tex
\begin{abstract}
Estimating robot pose from a monocular RGB image is a challenge in robotics and computer vision. Existing methods typically build networks on top of 2D visual backbones and depend heavily on labeled data for training, which is often scarce in real-world scenarios, causing a sim-to-real gap.
Moreover, these approaches reduce the 3D-based problem to 2D domain, neglecting the 3D priors. 
To address these, we propose Robot Topological Alignment Graph (RoboTAG), which incorporates a 3D branch to inject 3D priors while enabling co-evolution of the 2D and 3D representations, alleviating the reliance on labels. 
Specifically, the RoboTAG consists of a 3D branch and a 2D branch, where nodes represent the states of the camera and robot system, and edges capture the dependencies between these variables or denote alignments between them. Closed loops are then defined in the graph, on which a consistency supervision across branches can be applied. 
Experimental results demonstrate that our method is effective across robot types, suggesting new possibilities of alleviating the data bottleneck in robotics.
\end{abstract}

%% file: sec/1_intro.tex
\section{Introduction}
\label{sec:intro}

Robot pose estimation from monocular RGB images is a fundamental task in robotics and computer vision. Establishing a direct mapping between visual signals and robot configurations enables critical downstream applications such as human-robot interaction~\cite{hci1,hci2,hci3} and multi-robot collaboration~\cite{multi_robot1,multi_robot2,multi_robot3}, while also opening new possibilities for automatic robot data annotation and the integration of vision foundation models in robotics, including diffusion models~\cite{video_diff_1,video_diff_2,video_diff_3} for long-term planning~\cite{drrobot,zhen2025tesseract}.

However, this task is challenging due to the abstractness of robot configurations~\cite{holistic}. Existing methods~\cite{robopose,holistic,goswami2025robopepp} typically build networks on top of 2D visual backbones and depend heavily on labeled data for training, which makes deployment in real-world scenarios difficult. Moreover, these approaches reduce the inherently 3D problem to the 2D domain, neglecting rich priors available in pre-trained 3D models and suffering from the inherent spatial ambiguity of 2D representations~\cite{marr2010vision}, thus limiting performance.

\begin{figure}[t]
\centering
\includegraphics[width=0.45\textwidth]{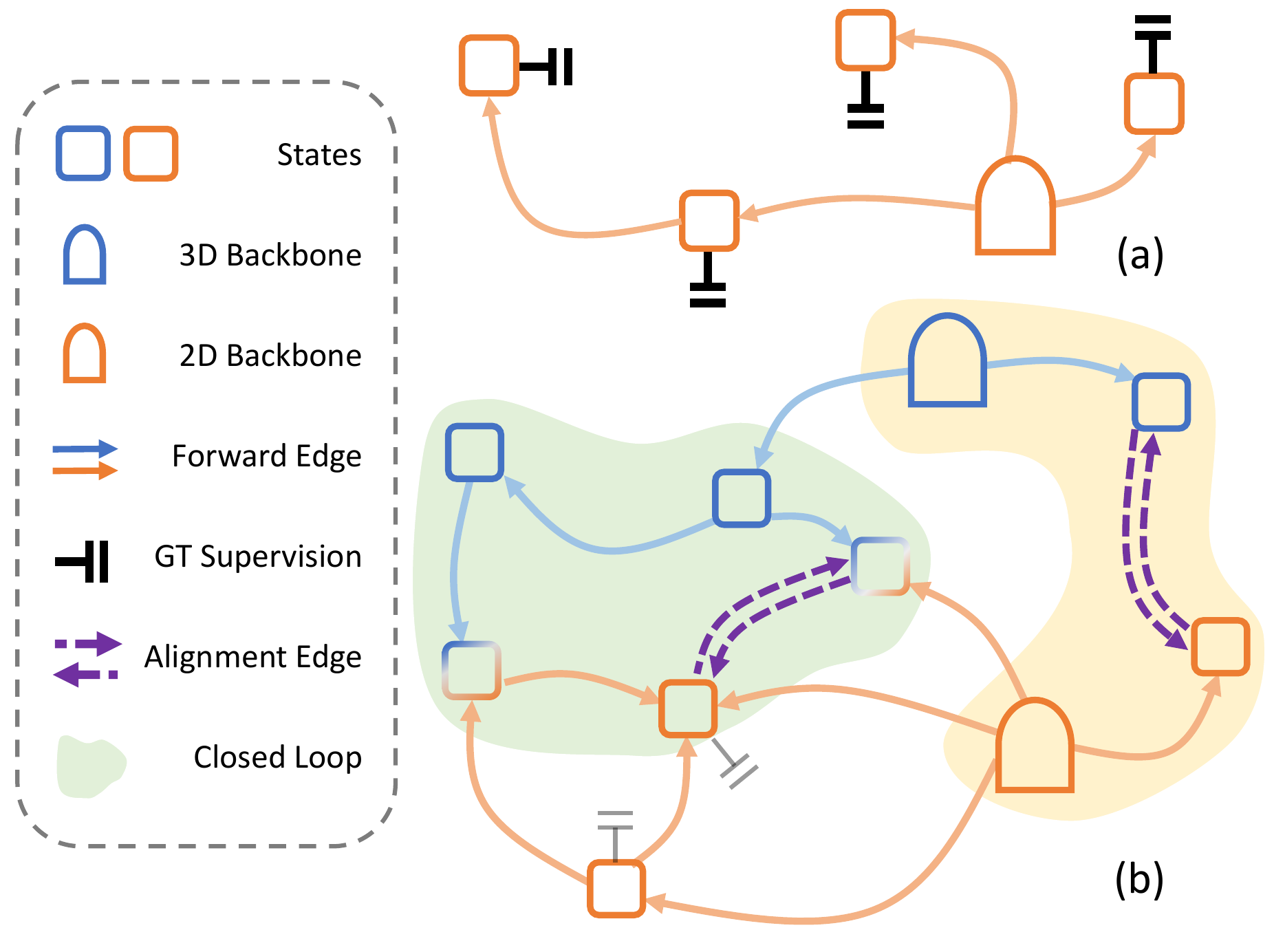} 
\caption{The intuition of RoboTAG. (a) Existing works predict every state (square) of the camera and robot system independently, and apply supervision on each of them. (b) Our method constructs a topological graph, RoboTAG, by introducing a 3D branch (blue) with forward edges (solid arrows) and alignment edges (dotted arrows) connecting system states. Multiple forward edges to a state indicate it depends on several others. Closed loops are defined in this topology to enable 2D-3D consistency supervision.}
\label{fig::header}
\end{figure}

To tackle these challenges, we propose \textbf{Robo}t \textbf{T}opological \textbf{A}lignment \textbf{G}raph (\textbf{RoboTAG}): 
A 3D prior is injected with a topological structure, which enables the co-evolution~\cite{liuff} of 2D and 3D representations, thereby alleviating reliance on labeled data. The name of RoboTAG reflects two aspects of our approach beyond abbreviation: (1) the framework acts as semantic tags for unannotated data through 2D-3D alignment, and (2) the end-to-end model can generate labels for in-the-wild robot videos.
Specifically, RoboTAG consists of dual branches: a 3D branch and a 2D branch, which constructs a graph in a unified system of a robot and cameras.
Nodes in the graph represent system state variables (camera extrinsics, robot joint angles, keypoints, etc.), while edges connecting these nodes are defined in two types:
(a) forward edges that capture dependencies between variables, and (b) alignment edges that denote correspondences between equivalent nodes across branches. As shown in Figure~\ref{fig::header}, these edges form a topology containing closed loops defined with specific constraints, enabling 2D-3D consistency supervision over the alignment edges. These closed loops strengthen the graph topology and enable alignment across branches.

With RoboTAG, the 2D and 3D branches are deeply intertwined, enabling co-evolution of both backbones. This design allows us to utilize in-the-wild images without annotations as an optional training stage. 
Extensive experiments across diverse robot embodiments and both real-world and simulated environments validate the effectiveness of our approach. This work suggests new possibilities for alleviating the data scarcity issue in robotics.

Our contributions are summarized as follows:
\begin{itemize}
    \item We introduce Robot Topological Alignment Graph (RoboTAG) for robot pose estimation, which effectively incorporates 3D priors and enables 2D-3D alignment.
    \item We propose a novel graph topology with closed loops that enables 2D-3D consistency supervision, allowing co-evolution of 2D and 3D representations.
    \item We achieve state-of-the-art performance across robot types and demonstrate the potential of mitigating the data bottleneck in robotics.
\end{itemize}

%% file: sec/2_related.tex
\section{Related Work}
\label{sec:related}
\textbf{Robot Hand-Eye Calibration.}
Traditional methods for robot hand-eye calibration~\cite{aruco,apriltag,artag} typically rely on a set of known correspondence points between the robot's end-effector and the camera. With known joint angles, the 3D positions of the correspondence points can be computed, and an optimization problem~\cite{hand2eye_eg1,hand2eye_eg2,hand2eye_eg3} is solved to find the camera-to-robot transformation. This, however, requires a set of known correspondence points in different robot configurations, which is hard to obtain in real-world scenarios.
Recent learning-based approach represented by~\cite{dream} tackles this problem by predicting the keypoints using networks and then doing camera pose estimation with Perspective-n-Points (PnP) algorithms~\cite{pnp0, pnp1,differentiable_pnp}. Further works like~\cite{posefusion,gsam,ctrnet,sgtapose} improve performance by introducing modern architectures or modules like feature fusion or temporal attention. 

\noindent\textbf{Robot Pose Estimation.}
Robot hand-eye calibration problems are often solved under the assumption of known robot joint angles, which limits the applicability of these methods in real-world settings like multi-robot collaboration. 
While significant progress for human pose estimation~\cite{zhu2024dpmesh,li2025scorehoi} has been made recently, works predicting robot joint angles are also presented.
RoboPose~\cite{robopose} proposed a refinement-based method that estimates the robot joint angles and camera pose by iteratively rendering and comparing, which requires significant computation. RoboKeyGen~\cite{robokeygen} utilizes a diffusion model to lift predicted 2D keypoints to 3D. RoboPEPP~\cite{goswami2025robopepp} further introduces a masked autoencoder (MAE)~\cite{mae} during training to give the network a robust understanding of the robot and camera system. These methods, however, tackle the problem in a 2D domain and do not leverage rich 3D information in this problem. Also, they rely heavily on labeled data for training, which is often not abundant in real-world deployments.

\begin{figure*}[t]
\centering
\includegraphics[width=0.95\textwidth]{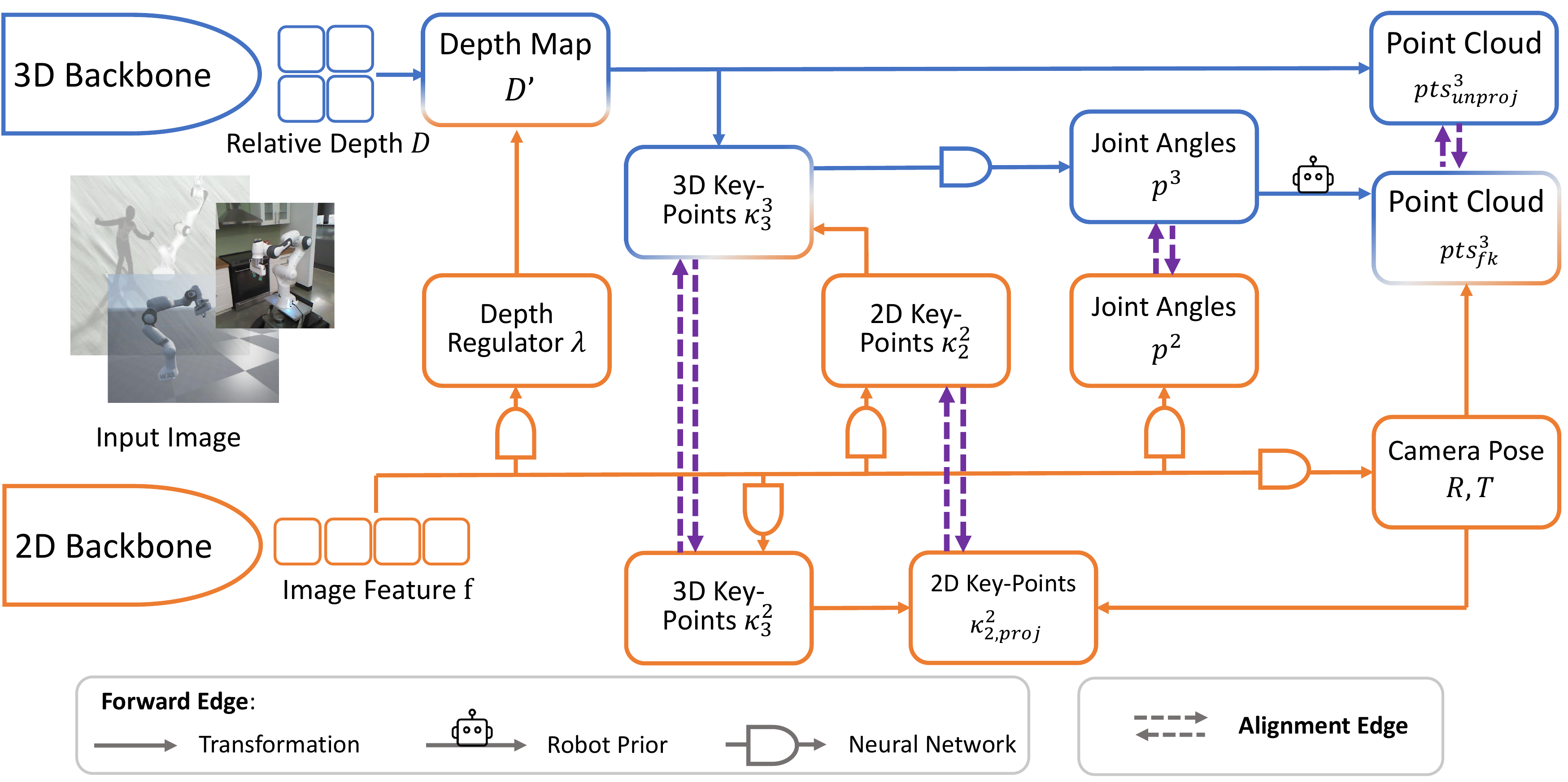} 
\caption{Overview of the proposed method. The framework consists of a 3D branch and a 2D branch, which are deeply intertwined as a topological graph. The nodes represent the states of the camera and robot system, and the edges represent dependencies between these variables (solid arrow) or denote alignments between equivalent nodes (dotted arrow). The closed loops in the topology enable 2D-3D consistency supervision, allowing the 2D and 3D branches to co-evolve. Gradients from 2D-3D alignment losses flow in the closed loops, enhancing the Neural Networks on the forward edges. Part of the graph, including the forward kinematics keypoints, is omitted for clarity.}
\label{fig::pipe}
\end{figure*}

%% file: sec/3_method.tex
\section{Method}
\label{sec:method}
\textbf{Problem Definition.} Given a monocular RGB image $I$ of an articulated robot with n joints, we define the camera and the robot as a system:
\begin{equation}
    \mathcal{S}_0=\{p, R, T\},
\end{equation}
where $p$ is the robot joint angles, or robot pose and robot configuration as used in literature interchangeably, $R$ and $T$ are the camera rotation and translation (referred as camera pose as a whole) relative to the robot base. These states, $p, R, T$ can fully determine the system $\mathcal{S}_0$. Our aim is to estimate the states of the system $\mathcal{S}_0$ from the image $I$, i.e., $\hat{\mathcal{S}}_0 = \{ \hat{p}(I), \hat{R}(I), \hat{T}(I) \}$, where $\hat{p}, \hat{R}, \hat{T}$ are the predicted robot joint angles, camera rotation and camera translation.

\noindent\textbf{Method Overview.}
Our presented framework, Robot Topological Alignment Graph (RoboTAG), consists of two branches, a 3D branch and a 2D branch, which are deeply intertwined as a topological graph. The nodes and edges of the graph are defined in Section~\ref{sec::define}. The construction method is introduced in Section~\ref{sec::graph-construction}. With this topology, we find a certain set of closed loops that satisfy certain constraints and 2D-3D consistency supervision for the closed loops in Section~\ref{sec::topo-constraints}, which enables co-evolution of the 3D and 2D branches. The whole framework can be trained end-to-end both in a supervised manner and in an aligning manner, which are detailed in Section~\ref{sec::training}.

\begin{algorithm}[tb]
\caption{Definition of TAG}
\label{alg::build}

\textbf{Input}: Image $I$\\
\textbf{Output}: Graph $G$, Topology $(X,\tau)$\\
\begin{algorithmic}[] 
\renewcommand{\COMMENT}[1]{\hfill$\triangleright$ #1}
\STATE Let $S_1=\{p, R, T, \kappa_2, \kappa_3, pts\}$.
\STATE Let $\{\mathcal{V}^{3}_n\}= \{D, p, \kappa_3, pts\}$.
\STATE Let $\{\mathcal{V}^{2}_n\}= \{f, p, R, T, \kappa_2, \kappa_3, pts\}$.
\STATE $\{\mathcal{V}_n\} = \{\mathcal{V}^{3}_n\} \cup \{\mathcal{V}^{2}_n\}$ \COMMENT{Define the set of nodes in $G$.}
\FOR{$\mathcal{V}_i, \mathcal{V}_j$ \textbf{ in } $\{\mathcal{V}_n\}$}
    \IF {$\text{if } \mathcal{V}_i \Longleftrightarrow \mathcal{V}_j$}
    \STATE $\mathcal{E}^{align}_{\mathcal{V}_i, \mathcal{V}_j}=1$ \COMMENT{Add edge to $G$.}
    \STATE $X \leftarrow X \cup \{\mathcal{E}^{align}_{\mathcal{V}_i, \mathcal{V}_j}\}$ \\
    \ELSIF {$\text{if } \frac{\partial \mathcal{V}_i}{\partial \mathcal{V}_j} \neq 0$}
    \STATE $\mathcal{E}^{forward}_{\mathcal{V}_i, \mathcal{V}_j}=1$ \COMMENT{Add edge to $G$.}
    \STATE $X \leftarrow X \cup \{\mathcal{E}^{forward}_{\mathcal{V}_i, \mathcal{V}_j}\}$ \\
    \ENDIF
\ENDFOR
\STATE $\mathcal{P} = \big\{ (e_1, e_2, \ldots, e_n) \mid e_k \in X,$
\STATE $\quad\quad\quad\text{and each } e_k \text{ connects } v_{k-1} \text{ to } v_k \text{ consecutively} \big\}$
\STATE $\tau = \left\{\, \bigcup_{i=1}^k \gamma_i \;\middle|\; \gamma_i \in \mathcal{P},\ k \in \mathbb{N}^{+} \right\}$. \\
\COMMENT{Topology $\tau$ is generated by $\mathcal{P}$ as a basis.}
\STATE \textbf{return} $X, \tau$
\end{algorithmic}
\end{algorithm}

\subsection{TAG Definition} \label{sec::define}

We define TAG in this section. Algorithm~\ref{alg::build} outlines the mathematical formulation of the graph and topology, and Figure~\ref{fig::pipe} illustrates the overall structure. 

\noindent\textbf{Dual-branch Nodes.} To construct the topological graph, we first define an expanded set of states for the system $\mathcal{S}_0$:
\begin{equation}
\mathcal{S}_1=\{p, R, T, \kappa_2, \kappa_3, pts\},
\end{equation}
where $\kappa_2$ and $\kappa_3$ are the 2D and 3D keypoints of the robot, and $pts$ is the 3D point cloud of the robot, all expressed in the root frame of the robot. As $\mathcal{S}_0$ already fully determines the system, $\mathcal{S}_1$ has redundant states where we can build our alignment supervision. We then define the 3D branch nodes $\{\mathcal{V}^{3}_n\}$ and 2D branch nodes $\{\mathcal{V}^{2}_n\}$ as:
\begin{equation}
    \{\mathcal{V}^{3}_n\}= \{D, D', p, \kappa_3, pts\},
\end{equation}
\begin{equation}
    \{\mathcal{V}^{2}_n\}= \{f, \lambda, p, R, T, \kappa_2, \kappa_3, pts\},
\end{equation}
where the $D$ and $D'$ denotes the relative depth estimated by the 3D backbone and the absolute depth respectively, $f$ is the image features extracted by 2D backbone, and $\lambda$ is the estimated depth regulator~\cite{holistic}. 
The nodes are denoted in Figure~\ref{fig::header} as the squares, and in Figure \ref{fig::pipe} as the named boxes.

\noindent\textbf{Forward Edges.}
We then define two types of edges in the graph connecting the nodes, forward edges and alignment edges. $\mathcal{E}_{\mathcal{V}_i, \mathcal{V}_j}$ denotes an edge between node $\mathcal{V}_i$ and $\mathcal{V}_j$. The forward edges capture the dependencies between the variables, demonstrated by the equation below:
\begin{equation}
    \mathcal{E}^{forward}_{\mathcal{V}_i, \mathcal{V}_j} = \left\{\begin{array}{ll}
        1, & \text{if } \frac{\partial \mathcal{V}_i}{\partial \mathcal{V}_j} \neq 0; \\
        0, & \text{else}. 
    \end{array}\right.
\end{equation}

For instance, robot joint angles $p$ and 3D keypoints $\kappa_3$ are dependent on each other, thus we have $\mathcal{E}^{forward}_{p, \kappa_3} = 1$; $p$ and camera rotation $R$ are not dependent on each other, thus we have $\mathcal{E}^{forward}_{p, R} = 0$. The forward edges are denoted in Figure \ref{fig::pipe} as the solid arrows. These dependencies are modeled by one of the following:
\begin{itemize}
    \item Transformation. If two states are mathematically equivalent or can be deduced from one another, we can directly perform a transformation. 
    \item Robot prior model. If two states are related by a robot prior model (URDF description), we can use the prior model to compute the relationship. 
    \item Neural network: Some states have implicit relationships, and we model these relationships with neural networks. 
\end{itemize}

Examples are provided in Supplementary.

\noindent\textbf{Alignment Edges.}
With the deliberated redundancy inside the nodes, we can define the alignment edges that denote the alignments between equivalent nodes across branches. The alignment edges are defined as: 
\begin{equation}
    \mathcal{E}^{align}_{\mathcal{V}_i, \mathcal{V}_j} = \left\{\begin{array}{ll}
        1, & \text{if } \mathcal{V}_i \Longleftrightarrow \mathcal{V}_j; \\
        0, & \text{else}. 
    \end{array}\right.
\end{equation}
The alignment edges are denoted in Figure~\ref{fig::header} and Figure \ref{fig::pipe} as the dashed arrows. Examples of the alignment edges could be found in Supplementary.

One topological basis $\mathcal{P}$ is defined as the set of all paths formed by the edges (both forward and alignment edges) in the graph, and the topology space is generated by $\mathcal{P}$. Any connected path composed of these edges is considered an element of the topology. Note that when defining paths, both forward and alignment edges are treated as undirected.

\subsection{TAG Construction} \label{sec::graph-construction}


This section outlines the construction of the TAG with the nodes and edges defined in Section~\ref{sec::define}. The construction process is illustrated in Figure~\ref{fig::pipe}.

The input image $I$ is processed with a 2D backbone to extract the image features $f$, and with a 3D backbone to extract the relative depth map $D$:
\begin{align}
    f &= \text{2D Backbone}(I),  \\
    D &= \text{3D Backbone}(I). 
\end{align}

Then, forward edges in 2D branch are constructed based on the dependencies of the nodes, as shown in Figure~\ref{fig::pipe}. Specifically, we have:
\begin{equation}
    \{p^2, R, T, \kappa_2^2, \kappa_3^2, pts^2, \lambda\} = \text{Forward }(f),
\end{equation}
where the subscript indicates the dimensionality of each variable, and the superscript denotes the branch (2D or 3D). $\lambda$ indicates the predicted depth regulator derived from image features, which is used to regulate the absolute value of depth~\cite{depth}.
The absolute depth map $D'$ from the predicted depthmap $D$ regulated by $\lambda$ following~\cite{holistic}: $D' = \lambda \cdot D$.
Forward edges are constructed to obtain 3D branch states:
\begin{equation}
    \{p^3, \kappa_2^3, \kappa_3^3, pts^3_{unproj}\} = \text{Forward }(D', R, T)
\end{equation}

With the joint angles $p^2, p^3$ and camera pose $R, T$, we can compute the 3D point cloud $pts^{2}_{fk}$, $pts^{3}_{fk}$ and the 3D keypoints $\kappa_{3,fk}^{2}$, $\kappa_{3,fk}^{3}$ in both the 2D branch and the 3D branch with robot forward kinematics, which introduces strong robot priors into the graph:
\begin{equation}
\scalebox{0.95}{$
    \{\kappa_{3,fk}^3, \kappa_{3,fk}^2, pts_{fk}^3, pts^2_{fk}\} = \text{Forward }(p^2, p^3, R, T)
$}.
\end{equation}

Projection from 3D to 2D can be performed to obtain the 2D keypoints from 3D keypoints:
\begin{equation}
    {\kappa_{2, proj}^3,\kappa_{2, proj}^2}  = \text{Forward }(\kappa_3^3, \kappa_3^2, R, T).
\end{equation}

Alignment edges are then constructed between the equivalent nodes across branches. Specifically, we have:
\begin{align} \label{eq::loops}
    p^2 &\Longleftrightarrow p^3, \\
    \kappa_3^2 &\Longleftrightarrow \kappa_3^3 \Longleftrightarrow \kappa_{3,fk}^3 \Longleftrightarrow \kappa_{3,fk}^2,\\
    \kappa_{2}^2 &\Longleftrightarrow \kappa_{2,proj}^2\Longleftrightarrow \kappa_{2,proj}^3, \\
    pts^{2}_{fk} &\Longleftrightarrow pts^{3}_{fk}\Longleftrightarrow pts^{3}_{unproj}. 
\end{align}

For a detailed procedure of building the graph step by step and a breakdown of the underlying relationships, we kindly refer readers to Supplementary.

\begin{algorithm}[tb]
\caption{Definition of Closed Loop in TAG}
\label{alg::loop}

\textbf{Input}: Topology $(X,\tau)$ \\
\textbf{Output}: Closed Loops $\mathcal{B}$\\
\begin{algorithmic}[] 
\renewcommand{\COMMENT}[1]{\hfill$\triangleright$ #1}
\STATE Let $L = \emptyset$ \COMMENT{Initialize the set of closed loops set.}
\STATE Let $Sp = \{D, f\}$ \COMMENT{Special states from 2D and 3D backbones.}
\FOR {$\gamma \in \tau$}
    \IF {$v_0 = v_n$ \textbf{ where } \\\quad \quad \quad \quad $\gamma = (v_0, e_1, v_1, e_2, \ldots, v_{n-1}, e_n, v_n)$}
        \IF { $\left| \left\{ e_k \in \gamma \mid e_k \text{ is an alignment edge} \right\} \right| = 1 $}
            \STATE $L \leftarrow L \cup \{\gamma\}$ \COMMENT{Add closed loop to $L$.}
        \ENDIF

    \ELSIF {$v_0, v_n \in Sp$ \textbf{ where } \\\quad \quad \quad \quad $\gamma = (v_0, e_1, v_1, e_2, \ldots, v_{n-1}, e_n, v_n)$}
        \IF { $\left| \left\{ e_k \in \gamma \mid e_k \text{ is an alignment edge} \right\} \right| = 1 $}
            \STATE $L \leftarrow L \cup \{\gamma\}$ \COMMENT{Add closed loop to $L$.}
        \ENDIF
    \ENDIF
\ENDFOR
\STATE Let $\mathcal{B} \subseteq L$ be a fundamental cycle basis such that 
\STATE \hspace{2em} $H_1(G, \mathbb{Z}_2) = \text{span}_{\mathbb{Z}_2}(\mathcal{B})$
\COMMENT Compute basis for $L$
\STATE \textbf{return} $\mathcal{B}$
\end{algorithmic}
\end{algorithm}

\begin{table*}[h]
    \centering
    \renewcommand{\arraystretch}{1.2}
    \label{table:main_comparison_auc_mean}
    \begin{tabular}{lccccccccccc} 
    \hline
    \multirow{2}{*}{Method} & \multirow{2}{*}{Average} 
    & Panda  & Panda  & Panda  & Panda 
    & Panda  & Panda  
    & Kuka  & Kuka 
    & Baxter \\
     &&3C-AK  &3C-XK  &3C-RS  &ORB   &DR  &Photo  &DR  &Photo   & DR   \\
    \hline
     *DREAM-F & -
    & 68.9 & 24.4 & 76.1 & 61.9 
    & 81.3 & 79.5 
    & - & - 
    & - \\
     *DREAM-Q & -
    & 52.4 & 37.5 & 78.0 & 57.1 
    & 77.8 & 74.3 
    & - & - 
    & 75.5 \\
     *DREAM-H & -
    & 60.5 & 64.0 & 78.8 & 69.1 
    & 82.9 & 81.1 
    & 73.3 & 72.1 
    & - \\
    \hline
    RoboPose & 71.3
    & 70.4 & \underline{77.6}  & 74.3 & 70.4 
    & \underline{82.9}  & 79.7 
    & \textbf{80.2} & 73.2 
    & 32.7 \\
    RoboPEPP & 74.0 
    & 75.3& \textbf{78.5} & \textbf{80.5}& \textbf{77.5}& \textbf{83.0}& \underline{84.1} & \underline{76.2} & \underline{76.1}& 34.4\\
    Holistic Pose& \underline{75.7} 
    & \underline{82.2}  & 76.0 & 75.2 & 75.2
    & 82.7 & 82.0 
    & 75.1 & 73.9 
    & \textbf{58.8} \\
    \hline
    Ours & \textbf{76.9}
    &\textbf{83.1} &75.7 &\underline{78.3}  &\textbf{77.5} &82.5 &\textbf{84.3} & 75.0& \textbf{76.6}& \textbf{58.8} \\
    \hline
    \end{tabular}
    \caption{Comparison of AUC $\uparrow$ of the ADD curve on DREAM datasets. Panda 3C datasets and Panda ORB are real-world photos, and the rest are synthetic datasets. Bold indicates the best result; underline indicates the second best. * denotes using ground-truth joint state parameters. We achieve SOTA on 5 out of 9 benchmarks, surpassing the second-best by 1.2\% in average.}
    \label{table:main_table}
\end{table*}

\subsection{Topological Closed Loops} \label{sec::topo-constraints}

\noindent 
Based on the forward edges and alignment edges constructed in Section~\ref{sec::graph-construction}, we further define two types of structures as ``Closed Loops" in the RoboTAG, which indicate the effective gradient flow during training. These closed loops are defined as follows:
A mathematical definition of the closed loops is outlined in Algorithm~\ref{alg::loop}. The intuitive definitions are as follows:
(1)\textbf{Alignment-Forward Loops.} These loops are enclosed by exactly one alignment edge and several forward edges. An example is shown in Figure~\ref{fig::header} in green.
(2)\textbf{Backbone-Connecting Lines.} These lines are not geometrically closed, but they start and end at one of the 3D backbone and 2D backbone, respectively. Exactly one edge is an alignment edge, and the rest are forward edges. An example is shown in Figure~\ref{fig::header} in yellow. Basically, the second type of structure can be viewed as a special case of the first type if we consider the 3D backbone and 2D backbone as two forward-edge-connected nodes. 

These defined closed loops constitute a topological basis for a subspace of the topological space, which is eligible for 2D-3D consistency supervision. For the alignment-forward loops, we can apply a contrastive loss between the two states connected by an alignment edge, guiding these two states to become consistent. The gradients from this loss will be backpropagated through the entire closed loop, training the neural networks that form the third type of forward edges. In the case of backbone-connecting lines, the gradients will be backpropagated into the 3D backbone and 2D backbone, enabling co-evolution of the two branches.

\subsection {Training} \label{sec::training}
The TAG can be trained in both supervised and alignment phases. 
In a supervised setting where some of the states have ground truth labels, we can apply both the supervised losses on these states. In an alignment setting where the data is only an unannotated input image, we only apply the 2D-3D consistency losses on the closed loops. The losses are defined as follows:
\begin{equation}
    \mathcal{L}_{\text{tot}} = \mathcal{L}_{\text{align}} + \mathcal{L}_{\text{supervised}}.
\end{equation}

2D-3D consistency loss $\mathcal{L}_{\text{align}}$ is defined as:
\begin{align}
    \mathcal{L}_{\text{align}} =\  &\frac{\alpha_1}{n} \|p^3 - p^{2}\|_2^2 \nonumber + \frac{\alpha_2}{m} \|\kappa_3^3 - \kappa_3^{2}\|_2^2 \nonumber \\
    &+ \frac{\alpha_3}{m} \|\kappa_3^{3} - \kappa_{3, fk}^{2}\|_2^2 \nonumber + \frac{\alpha_4}{m}\|\kappa_{2,proj}^{2} - \kappa_2^{2}\|_2^2 \nonumber \\
    &+ \alpha_5\cdot\mathcal{L}_{\text{Chamfer}}^{\text{uni}}(pts^3_{unproj}, pts^3_{fk}) \nonumber \\ &+ \alpha_6\cdot\mathcal{L}_{\text{Chamfer}}(pts^3_{unproj}, pts^2_{fk}) ,
\label{eq:align_loss}
\end{align}
where $n$ is the number of joints and $m$ is the number of keypoints, ``uni" indicates unidirectional Chamfer Distance. $\alpha_i$ are the hyperparameters to balance the losses. 

The supervised loss $\mathcal{L}_{\text{supervised}}$ is defined as:
\begin{align}
    \mathcal{L}_{\text{supervised}} =\  &\frac{\beta_1}{n} \|p^3 - p^{gt}\|_2^2 \nonumber \\ &+ \beta_2 \|R^2 - R^{gt}\|_2^2 \nonumber + \beta_3 \|T^2 - T^{gt}\|_2^2 \nonumber \\
    &+ \frac{\beta_4}{m} \|\kappa_2^{2} - \kappa_2^{gt}\|_2^2 \nonumber + \frac{\beta_5}{m} \|\kappa_{2,proj}^{2} - \kappa_2^{gt}\|_2^2 \nonumber \\
    &+ \beta_6\cdot\mathcal{L}_{\text{Chamfer}}^{\text{uni}}(pts^3_{unproj}, pts^{gt}_{fk}) \nonumber \\ &+ \beta_7\cdot\mathcal{L}_{\text{Chamfer}}(pts^2_{fk}, pts^{gt}_{fk}) ,
\label{eq:gt_loss}
\end{align}
where superscript $gt$ means ground truth labels, and 
$\beta_i$ are the hyperparameters to balance the losses. If some of the ground truth labels are not provided, the corresponding term is not used in the loss.

%% file: sec/4_experiments.tex
\begin{figure*}[t]
\centering
\includegraphics[width=0.95\textwidth]{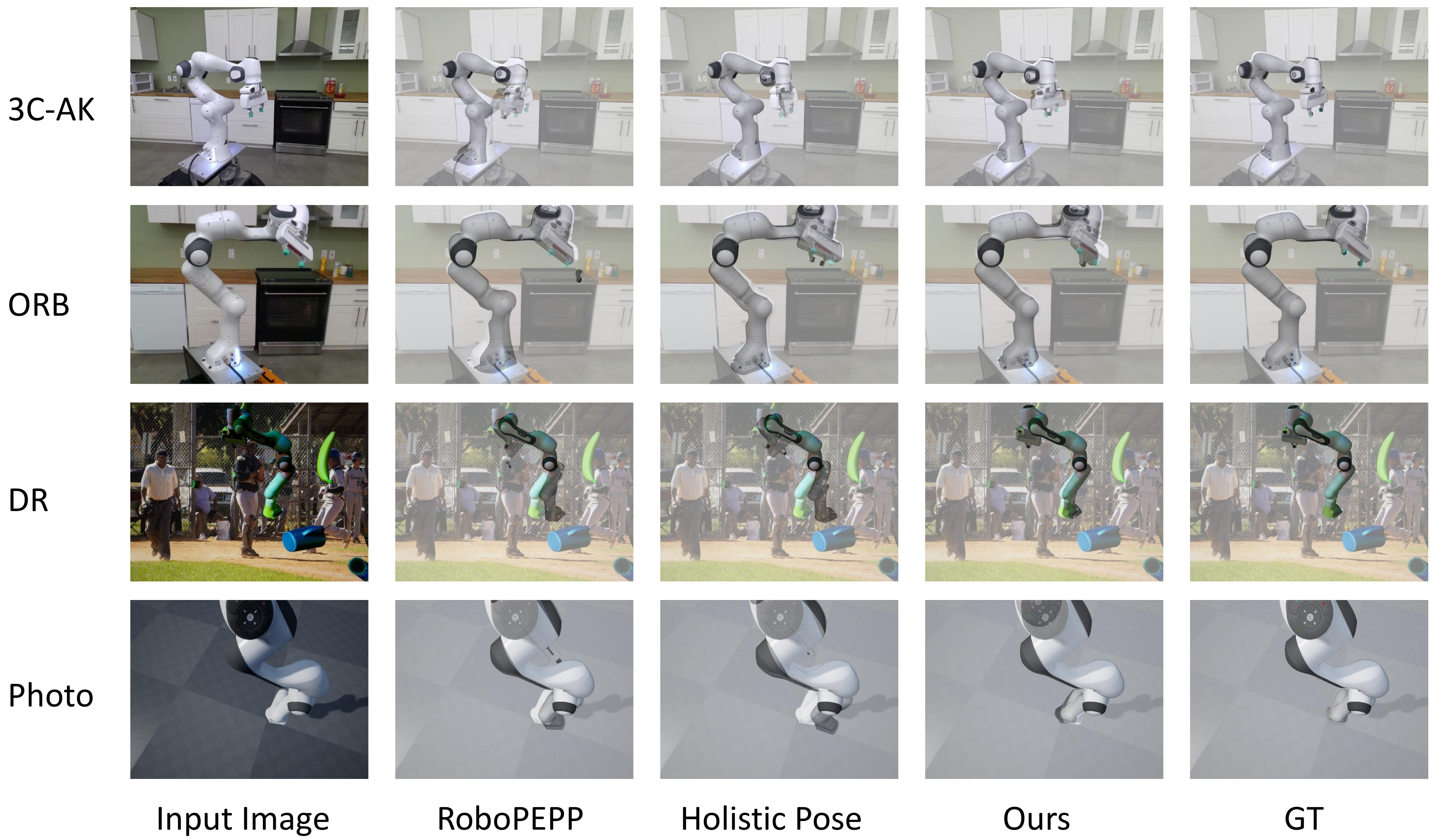} 
\caption{Qualitative results on the Panda real and synthetic datasets in DREAM. The predicted robot pose is overlaid on top of the input image. The smaller the gap between the grey render and the original robot is, the better the prediction is. Our method achieves the best performance for most robot parts.}
\label{fig::qualitative}
\end{figure*}

\section{Experiments}
\label{sec:experiments}
\subsection{Implementation Details}
Same to~\cite{robopose,goswami2025robopepp,holistic}, we train and evaluate the method on DREAM~\cite{dream}. For the three robots (Panda, Kuka, and Baxter), we train on the synthetic DR-train split. For Panda, the model is evaluated on two synthetic sets, DR and Photo, and four real-world sets (3C-AK, 3C-XK, 3C-RS, ORB); for Kuka, the eval set is DR and Photo; for Baxter, the eval set is DR. 
As the submodules of our TAG, the 2D backbone is initialized from the rootnet of~\cite{holistic} trained on the DepthNet task, and the 3D backbone is the pre-trained DepthAnything-V2~\cite{depth_anything_v2} Tiny. All of the weights are trainable during our training phases. 

We train our model on 8 NVIDIA L40S GPUs with 46GB memory each, and training converges in 3 days. We train the model in a hybrid manner with both cross-dimensional consistency losses and supervised losses on the annotated DR train sets. The learning rate is set to $1.2e^{-4}$ with a global batch size of 128, and the model is trained for 80 epochs with a decay rate of 0.95. For cases where real-world images are available (e.g. Panda 3C), we then train the model with only cross-dimensional consistency losses without labels. No other annotations or new data are introduced. The learning rate is set to $1.0e^{-6}$ with a batch size of 128, and the model is trained for 20 epochs with a decay rate of 0.95. More details can be found in the Supplementary Material.

\subsection{Results}
\textbf{Main Performance.} To the best of our knowledge, RoboTAG is the first approach to leverage pre-trained 3D backbones for robot pose estimation. As no prior work explores this direction, we do not have direct baselines for this component. We compare our method, RoboTAG, with three state-of-the-art approaches: RoboPose~\cite{robopose}, RoboPEPP~\cite{goswami2025robopepp}, and Holistic Pose~\cite{holistic}. For reference, we also list results from DREAM~\cite{dream}, where joint angles are provided.
RoboKeyGen~\cite{robokeygen} is not included in our comparison because it does not evaluate on commonly used benchmarks such as DREAM~\cite{dream}, relying instead on its own dataset. Moreover, the project has not been open-sourced.

We evaluate the models with the ADD metric between the predicted and ground truth joint positions inside the image following~\citet{goswami2025robopepp}. We report the Area Under Curve (AUC) of the ADD distribution curve in Table~\ref{table:main_table}. A higher AUC indicates better accuracy. 
As shown in Table~\ref{table:main_table}, RoboTAG achieves the highest average AUC of 76.9\%, outperforming all other methods by a margin of 1.2\%. Notably, our method achieves strong results on both real-world datasets (e.g., Panda 3C-AK) and synthetic photorealistic datasets (the ``Photo" sets). 
Since the supervised training set is DR, the other evaluation sets are out-of-distribution (OOD) relative to the DR evaluation set. RoboTAG's strong performance on these OOD sets highlights its robust generalization, which is important for real-world deployment.

DREAM~\cite{dream} and RoboPEPP~\cite{goswami2025robopepp} rely on the PnP~\cite{differentiable_pnp} algorithm to solve the camera pose based on the predicted 2D keypoints; as a result, the noise in the 2D predicted keypoints inevitably affects their camera. For instance, if there are points occluded or missing in the photo, PnP will suffer. For RoboPEPP, the deviation in 3D keypoints from forward kinematics with the predicted joint angles further aggravates the problem due to the accumulated error from joint pose estimation. In contrast, we use two practices to avoid this issue. On one hand, we directly predict the camera pose from our 2D branch, skipping the PnP step; on the other hand, camera poses are connected inside several closed loops, which supervise it in various directions, making it robust for different tasks. Holistic Pose~\cite{holistic} uses a 2D backbone to predict the 3D keypoints, which is a rather challenging task for the prediction head, while ours sets a closed loop on this task with the guidance of a 3D backbone, leading to a more accurate prediction.
We attribute the lower performance compared with the supervised method on the DR set to its close resemblance to the training data, making the hybrid framework with alignment loss less effective than direct supervised training. Nevertheless, we argue that the gains in the OOD sets, which are more important for real-world deployment, and the advantage in average AUC justify the effectiveness of our method.

\noindent\textbf{Latency.} RoboTAG runs in a single forward pass with lightweight backbones; inference takes 35 ms—close to 2D baselines (23 ms, 27 ms) and far below optimization-based ones (200 ms+).

\noindent\textbf{Qualitative Results.}
Qualitative results are shown in Figure~\ref{fig::qualitative}, where the predicted robot pose is overlaid on top of the input image using the render in grey. The smaller the gap between the render and the original shape is, the better the prediction is. 
In the visualization, our method can accurately estimate the robot joint angles and 3D keypoints in both the real world and synthetic scenarios, effectively covering the white mask.
In the noisy background in DR dataset and extreme views in Photo dataset (lines 3 and 4), our method exhibits a more robust performance compared with other methods. 
While 2D-based methods, RoboPEPP and Holistic Pose, struggle with the noisy features or extreme views that are out of distribution, 
our model can still constrain the basic 3D geometry with the 3D model, preventing unreasonable predictions. We attribute this to the closed loops in TAG. More visualizations can be found in the Supplementary.

Notably, our method does better on the parts close to the base (e.g., first link), while the parts far from the base (e.g,. gripper) have larger deviations. We attribute this as the joints near the base having a larger influence on 3D positions, which are thus more strictly supervised by our 3D branch. With the motion of the last joint, the 3D keypoints correspond weakly, demonstrated by the minor changes in 3D features in the 3D backbone, and are hard to constrain.

\begin{table}[h]
\centering
\renewcommand{\arraystretch}{1.1}
\setlength{\tabcolsep}{2.5pt}
\begin{tabular}{l l l l l l l l l}
  \hline
  Dataset& Method & J1 & J2 & J3 & J4 & J5 & J6 & Avg. \\
  \hline
  \multirow{4}{*}{\makecell{Panda\\DR}} & RoboPose & 6.1& 2.7& 3.6& 2.5& 6.3& 8.1& 4.9\\
   & RoboPEPP & 4.9& 2.3& 2.7& 2.2& 4.9&\textbf{5.4}& 3.8\\
   & Holistic Pose& 6.2&2.2& 3.9&1.9& 5.9& 6.6&  4.4\\
   & \textbf{Ours} & \textbf{4.7} & \textbf{2.0} & \textbf{2.5} & \textbf{1.7} & \textbf{4.7} & 6.2 & \textbf{3.6}\\
   
  \hline
 \multirow{4}{*}{\makecell{Panda\\Photo}} & RoboPose & 7.7& 3.5& 4.3& 3.4& 7.3& 8.1& 5.7\\
   & RoboPEPP & 4.4& \textbf{1.8}&\textbf{2.2}& 1.8& 4.4& \textbf{4.8}&\textbf{3.2}\\
   & Holistic Pose& 6.1& 2.2& 3.6& 2.0& 6.2& 6.6&   4.5\\
   & \textbf{Ours} & \textbf{3.7} & \textbf{1.8} & 2.4 & \textbf{1.7} & \textbf{4.1} & 6.3 & 3.3\\
   \hline
   
\end{tabular}
\caption{Mean Joint Angle Deviation in degrees $\downarrow$ on the Panda datasets in DREAM. Bold indicates the best result. We achieve SOTA performance for most of the joint angles which have significant influence on robot posture.}
\label{tab:joints}
\end{table}

\noindent\textbf{Mean Joint Angle Deviation.} Following~\cite{goswami2025robopepp}, we report the average of the absolute difference between the predicted and ground truth joint angles. A smaller value indicates better control accuracy. It can be seen in Table~\ref{tab:joints} that our method achieves strong performance on the first five joints, getting 4 or 5 SOTAs on each dataset, achieving the best average performance on the Panda DR dataset and comparable performance with RoboPEPP on the Panda Photo dataset. The sixth joint, which is farthest away from the base, has a larger deviation compared with RoboPEPP. This observation is aligned with the qualitative results in Figure~\ref{fig::qualitative}. Notably, the last joint has a much smaller impact on the global posture of the robot arm and the position of the end effector; thus, the deviation is less critical in real-world applications. Nevertheless, methods to further improve the accuracy of the last joint are worth exploring in future work, as introduced in the limitations.

\subsection{Ablation Study}
\begin{figure}[t]
\centering
\includegraphics[width=0.48\textwidth]{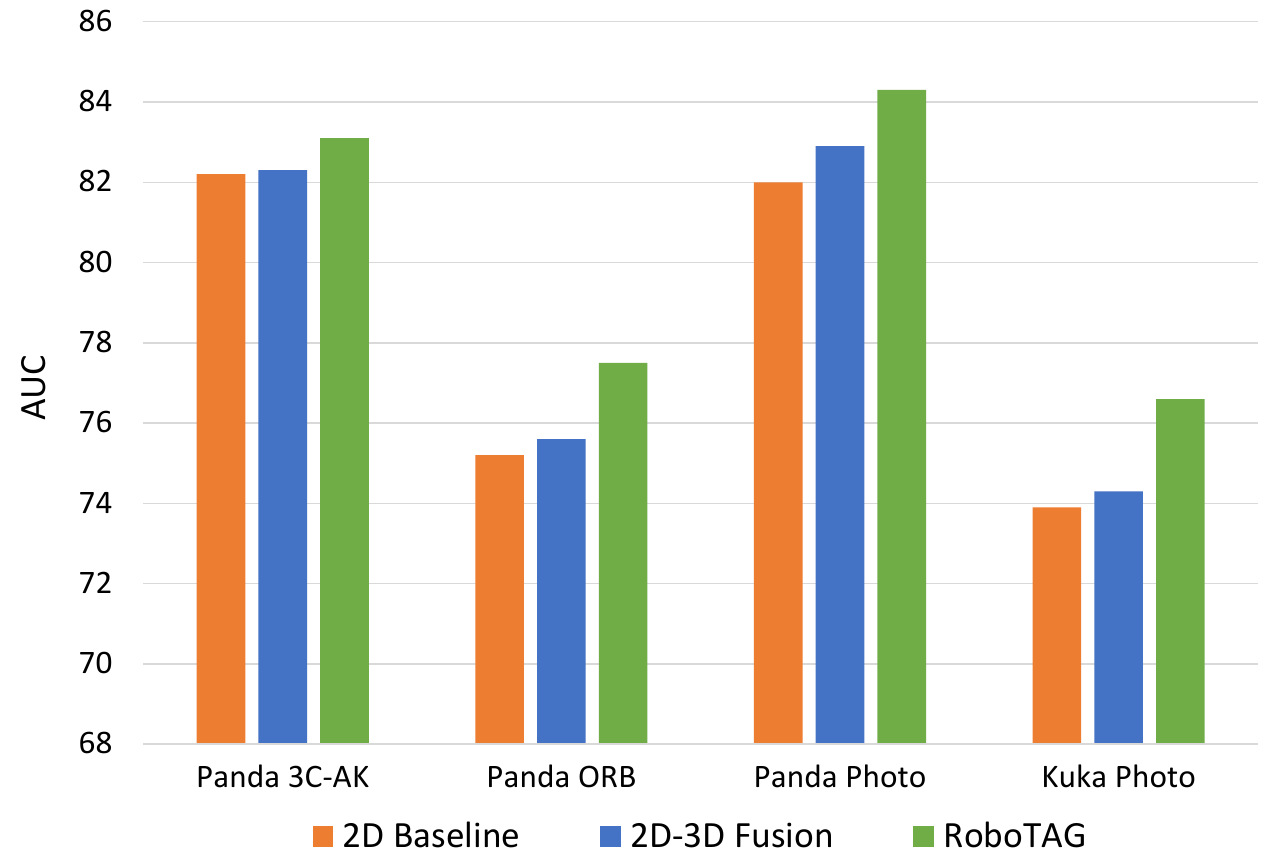} 
\caption{Ablation study for the effectiveness of 3D priors and TAG alignment on the Panda and Kuka datasets. Introducing 3D priors by adding 2D-3D feature fusion leads to a small performance gain of 0.45\%, and TAG further brings a boost of 1.6\%, demonstrating the effectiveness of our method.}
\label{fig::ablation}
\end{figure}

We evaluate the effectiveness of the 3D priors and the TAG alignment method on the real and synthetic datasets of Panda and Kuka. The results are shown in Figure~\ref{fig::ablation}. 

\noindent\textbf{3D Priors.} Theoretically, robot pose estimation involve complex 3D geometry information, which benefits from 3D priors. To compare the effectiveness of 3D priors, we set up a minimal comparison with the 2D branch only~\cite{holistic} and fusing the 3D prior. 

\noindent (1) 2D-only. We use the pre-trained backbone (ResNet50~\cite{resnet}) and prediction heads from~\cite{holistic} to directly predict $S_0$. This variant does not include any 3D branch or TAG structure, serving as the 2D baseline.

\noindent(2) 2D-3D fusion. We use the same 2D backbone and prediction heads as in the 2D-only baseline, but add a 3D backbone (DepthAnythingV2~\cite{depth_anything_v2}) to extract 3D features. The last layers of the 2D and 3D backbone are fused into a hybrid feature map using Hiwin attention~\cite{llavauhdv2}, a locality-preserving feature fusion method. The fused feature map is then used to predict the states via prediction heads adopted from~\cite{holistic}. No graph structure or alignment losses are used in this variant.
Experiment results show that with the 3D features, the performance improved on the four datasets (0.45\%), indicating the effectiveness of the 3D priors. 

\noindent\textbf{TAG.} We further ablate the effectiveness of the alignment with TAG. We compare the performance of our model of full TAG structure, with the closed loops and the corresponding 2D-3D consistency loss, against the feature fusion with Hiwin attention variant. The results demonstrate a significant further performance improvement (1.6\%) on the four datasets. We attribute this improvement to two points: (1) the decentralized structure of TAG. In the fusion variant, the fusion process is causing an information bottleneck, which limits the model's potential to exploit the 2D and 3D features, causing suboptimal performance. In contrast, the TAG structure is decentralized and thus no information bottleneck is introduced. (2) multi-target optimization. The 2D-3D consistency loss applied to the closed loops in TAG enables the networks on the relevant edges to be trained for different tasks, including dense predictions like point cloud, which is more robust than solely depending on sparse GT supervision. These two advantages of TAG bring about the improved leverage of 3D features over direct feature fusion. 

\begin{table}[ht]
    \centering
    \renewcommand{\arraystretch}{1.2}
    \setlength{\tabcolsep}{3pt}
    \begin{tabular}{lcccc} 
    \hline
    \multirow{2}{*}{Method} & Panda & Panda & Panda & Kuka \\
     & 3C-AK & ORB & Photo & Photo \\
    \hline
    Baseline &  82.2& 75.2 & 82.0 & 73.9 \\
    + Keypoints & 82.8\growth{+0.6} & 76.0\growth{+0.8} & 83.8\growth{+1.8} & 75.6\growth{+1.7} \\
    + Joints & 83.0\growth{+0.2} & 77.1\growth{+1.1} & 84.4\growth{+0.6} & 76.3\growth{+0.7} \\
    + PC (Ours) & 83.1\growth{+0.1} & 77.5\growth{+0.4} & 84.3\growth{-0.1} & 76.6\growth{+0.3} \\
    \hline
    \end{tabular}
    \caption{Ablation on the effectiveness of different types of closed loops in TAG. The number in the bracket is the increase compared with the previous row. Each type of loop has shown performance gain, and the final model with all loops achieves the best performance.}
    \label{table:loop_ablation}
    \vspace{-0.5em}
\end{table}

\noindent\textbf{Closed Loops.} We ablate the effectiveness of each type of closed loop in TAG listed in Equations 14-17 in Table.~\ref{table:loop_ablation}. The baseline is the model without any loops. Each row adds a new type of loop to the previous row. We classify the closed loops in TAG with the alignment edge it contains: 2D and 3D keypoints, joints, and point cloud (PC). We add them to the base model one by one and ablate the results across different robot types.

From this table, we see improvements across all types of closed loops. The 2D and 3D keypoints loop provides the most significant improvement, which is because the 2D and 3D keypoints are strong geometry priors that has loops containing vital networks 2D and 3D backbones, camera head, and depth regulator. Joint angles also provide significant improvement. We attribute this to the decisive importance of joint angles themselves, as they fully determine the relative robot posture. Alignment loops for point clouds also provide a performance gain, and we attribute this to the dense supervision signal it provides for the 3D backbone. All of our proposed loops are effective, and the final model with all loops achieves the best performance.

%% file: sec/5_conclusion.tex
\section{Conclusion}
\label{sec:conclusion}
In summary, we present RoboTAG, a novel topological graph framework that models the relationships between camera and robot states from both 2D and 3D perspectives. By leveraging forward and alignment edges as a topological basis, RoboTAG enables the definition of closed loops for 2D-3D consistency supervision, allowing the model to benefit from both 2D and 3D features. Extensive experiments demonstrate that RoboTAG achieves state-of-the-art performance across robot types, outperforming existing methods that rely solely on 2D information and annotated training data. Ablation studies further validate the effectiveness of 3D priors and the TAG structure in enhancing robot pose estimation. RoboTAG suggests new possibilities for addressing data scarcity in robotics.

\noindent\textbf{Limitations and Future Work.} 
We observed gaps in the experiments that could be addressed by future work. (1) Though TAG exhibits superior generalization, performance on the in-distribution DR set is slightly lower than existing methods, which suggests the 2D-3D consistency loss is less effective for in-distribution fitting when supervision signal is available. The DR drop reflects forgetting, a known issue with heterogeneous data. A unified design for the network architecture or training curricula might be valuable. (2) Though the accuracy for joint angle prediction and the global 3D posture is improved, the joints near the gripper, whose impact on the 3D structure is weak, still have larger deviations. Finer-grained perception of the gripper parts or hierarchical networks might be a promising direction.

\section*{Acknowledgements}
This work was partially supported by NIH grant R01HD104969, NIH grant 1U01CA284207, NSF grant CRCNS-2309041, and NTU Nanyang Assistant Professorship Startup Grant 025661-00012.

%% file: sec/X_suppl.tex
\clearpage
\setcounter{page}{1}
\maketitlesupplementary

\section{Method Details}
\subsection{Examples of Edges in TAG}
\textbf{Forward Edges.} In the TAG, we define forward edges as the dependencies between states where one state relies on the other.
For instance, robot joint angles $p$ and 3D keypoints $\kappa_3$ are dependent, because when robot joints change, the position of 3D keypoints $\kappa_3$ will also change. Thus we have $\mathcal{E}^{forward}_{p, \kappa_3} = 1$. $p$ and camera rotation $R$ are not dependent on each other, thus we have $\mathcal{E}^{forward}_{p, R} = 0$. The forward edges are denoted in Figure 1 and Figure 2 as the solid arrows. These dependencies are modeled by either one of the following, each provided with an example. For simplicity, we omit some of the superscripts denoting the branch.
\begin{itemize}
    \item Transformation. If two states are mathematically equivalent or can be deduced from one another, we can directly perform a transformation. 
    For instance, $R, T, \kappa_3$ can deduce $\kappa_2$ with a simple camera projection. We denote these edges with plain solid arrows in Figure 2.
    \item Robot prior model. If two states are related by a robot prior model, we can use the prior model to compute the relationship. 
    For instance, $p$ and $pts_{fk}$ are related by the robot URDF mesh, thus we can compute $pts_{fk}$ from $p$ with forward kinematics. We denote these edges with arrows with a robot icon.
    \item Neural network: Some states have implicit relationships, and we denote these relationships with neural networks. 
    For instance, $f$ and $p^2$ are related by a neural network that predicts the robot joint angles from the image features. We denote these edges with arrows with a half-circle shape.
\end{itemize}

\noindent\textbf{Alignment Edges.}
With the deliberated redundancy inside the nodes, we can define the alignment edges that denote the alignments between equivalent nodes across branches.
The alignment edges are denoted in Figure 1 and Figure 2 as dashed arrows.
Typical examples of alignment edges are:
\begin{itemize}
    \item 3D keypoints. The 3D keypoints $\kappa_3$ from 2D and 3D branches are equivalent, thus we have $\mathcal{E}^{align}_{\kappa_3^3, \kappa_3^2} = 1$.
    \item Robot joint angles. The robot joint angles $p$ from direct 2D prediction and inferred from 3D keypoints are equivalent, thus we have $\mathcal{E}^{align}_{p^3, p^2} = 1$.
\end{itemize}

\subsection{Details of Building TAG}
In this section, we provide detailed steps for building the Topological Alignment Graph (TAG) for our method. After defining the nodes and edges in Section 3.1, the construction of the graph can be described as follows. 

We first process the input image $I$ with a 2D backbone to extract the image features $f$, and with a 3D backbone to extract the depth map $D$. 

\begin{align}
    f &= \text{2D Backbone}(I),  \\
    D &= \text{3D Backbone}(I). 
\end{align}

In the 2D branch, we use separate networks to predict the robot joint angles $p$, camera rotation $R$, camera translation $T$, 2D keypoints $\kappa_2$, and 3D keypoints $\kappa_3$ from the image features $f$ in a feed-forward manner. 

\begin{align}
    p^2 &= \text{MLP}(f),  \\
    R &= \text{MLP}(f),  \\
    T &= \text{MLP}(f),  \\
    \lambda &= \text{MLP}(f),  \\
    \kappa_2^2 &= \text{Convolution}(f),  \\
    \kappa_3^2 &= \text{Convolution}(f),
\end{align}
where $\lambda$ is the predicted depth regulator derived from image features. According to the definition, these nodes all have forward edges from the 2D backbone image features $f$. 

Due to the relative nature of depth estimation~\cite{depth}, we obtain the absolute depth map $D'$ from the predicted depthmap $D$ regulated by $\lambda$ following~\cite{holistic}:

\begin{equation}
    D' = \lambda \cdot D.
\end{equation}

This forms a forward edge from the depth regulator $\lambda$ to the depth map $D'$. 

3D keypoints $\kappa_3^3$ can be unprojected from the depth map $D'$ with the 2D keypoints $\kappa_2^2$ into the camera space, which forms a forward edge from the 2D keypoints $\kappa_2^2$ and depth map $D'$ to the 3D keypoints $\kappa_3^3$:
\begin{equation}
    \kappa_3^3 = \text{Unproject}(D', \kappa_2^2).
\end{equation}

The 3D keypoints from the 3D branch $\kappa_3^3$ are equivalent to the 3D keypoints from the 2D branch $\kappa_3^2$; thus, we add an alignment edge between them.

3D keypoints have strong geometrical relationships with the robot joint angles $p$, thus we leverage a small network to predict the robot joint angles $p^3$ from the 3D keypoints $\kappa_3^3$:
\begin{equation}
    p^3 = \text{MLP}(\kappa_3^3).
\end{equation}

This equation forms a forward edge from the 3D keypoints $\kappa_3^3$ to the robot joint angles $p^3$, which can further form an alignment edge with the robot joint angles $p^2$ from the 2D branch.

With the robot mesh prior, joint angles $p$ can be used to compute the 3D point cloud $pts$ as well as the 3D keypoints $\kappa_3$ in both the 3D branch and the 2D branch. We denote the 3D point cloud as $pts^{2,3}_{fk}$ and the 3D keypoints as $\kappa_{3,fk}^{2,3}$ in 2D and 3D branch respectively. Please note that for simplicity, this part of the computation is not shown in Figure 2. The computation is done with forward kinematics:
\begin{align}
    pts^{2}_{fk} &= \text{ForwardKinematics}(p^2), \\
    pts^{3}_{fk} &= \text{ForwardKinematics}(p^3), \\
    \kappa_{3,fk}^{2} &= \text{ForwardKinematics}(\kappa_2^2), \\
    \kappa_{3,fk}^{3} &= \text{ForwardKinematics}(\kappa_3^3).
\end{align}

By building these forward edges, the 2D and 3D branches have been deeply intertwined. The corresponding 2D and 3D forward kinematics keypoints and point clouds are equivalent, we add alignment edges between them. 

Notably, two extra alignment edges are built to enrich the topology of the graph: (1) the 3D forward kinematics pointcloud should contain the 3D unprojected pointcloud, thus we add an alignment edge between the 3D forward kinematics pointcloud $pts^{3}_{fk}$ and the unprojected pointcloud $pts^{3}_{unproj}$, which is equivalent to the depth map $D'$ by definition; 
\begin{equation}
    pts^{3}_{unproj} = \text{Unproject}(D').
\end{equation}
(2) The 2D forward kinematics keypoints are equivalent to the 3D unprojected keypoints, thus we add an alignment edge between the 2D forward kinematics keypoints $\kappa_{2,fk}^{2}$ and the 3D unprojected keypoints $\kappa_3^{3}$. Thus, we get the final set of nodes and edges in the TAG, as shown in Figure 2. For brevity, some of the forward kinematics nodes are not shown in the figure.

\section{Training}
We acknowledge that the loss weights influence both accuracy and training stability. In practice, we performed a log-scale ablation and selected an empirically validated and stable configuration: $[\frac{\alpha_1}{n},...\alpha_6]=[1,1,1,1,0.1,0.1]$ and $[\frac{\beta_1}{n},...\beta_7]=[0.1,0.1,0.1,1,1,0.1,0.1]$.

In the alignment stage, we use several designs to avoid collapsed or trivial convergence: (1) pre-trained 2D and 3D backbones provide strong non-degenerate priors; (2) geometric constraints from robot FK enforce physically meaningful predictions; and (3) controlled optimization (e.g., warm up) stabilizes training.

\section{More Qualitative Results}
We provide more qualitative comparison on our method and the current SOTA Holistic Pose~\cite{holistic} in Figure~\ref{fig::more_viz}. 
For visualization, we use the predicted robot joint angles to configure the URDF robot mesh into the corresponding target posture, and then render the mesh with the camera poses onto the image plane. The rendered robot arm is in grey color, which is then blended with the input image.
The smaller the margin between the grey rendered shape and the original robot in the input image is, the better the prediction is. Our method is able to predict the robot configuration accurately, even in challenging scenarios such as clustered or noisy backgrounds, partially out-of-view robots, and uncommon view points. This demonstrates the potential of our method in real-world applications.

\setcounter{figure}{0}
\renewcommand{\thefigure}{A}
\begin{figure*}[t]
\centering
\includegraphics[width=0.95\textwidth]{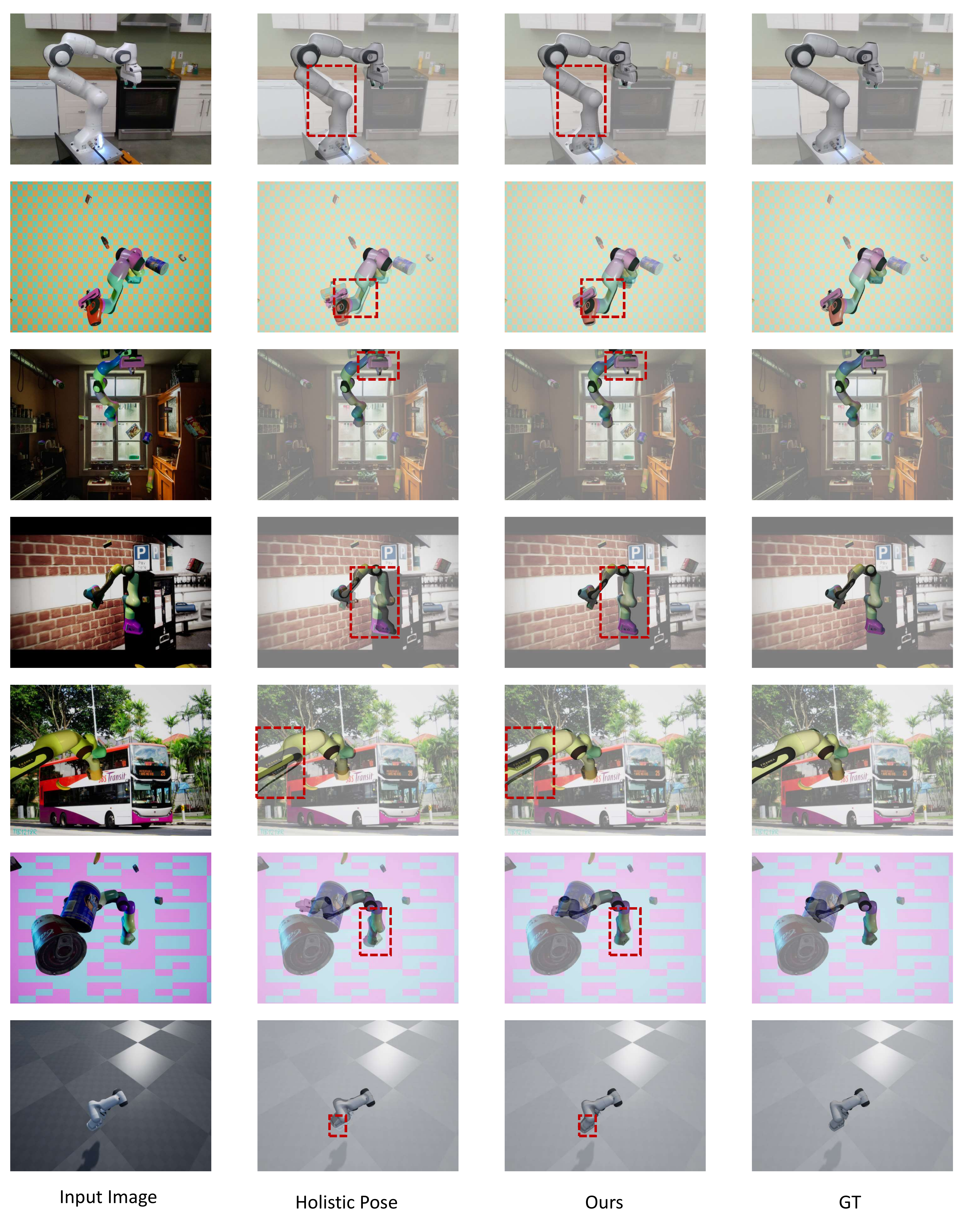} 
\caption{More visualization on our method and current SOTA Holistic Pose~\cite{holistic}. Significant differences are highlighted in red boxes. }
\label{fig::more_viz}
\end{figure*}